%% file: main.tex
\documentclass[10pt,twocolumn,letterpaper]{article}

\usepackage[pagenumbers]{cvpr} 
\usepackage{wrapfig}

\input{preamble}

\definecolor{cvprblue}{rgb}{0.21,0.49,0.74}

\def\paperID{5707} 
\def\confName{CVPR}
\def\confYear{2025}

\title{\algo: Mixed Precision Quantization for Diffusion Models}

\author{
Rocco Manz Maruzzelli} 

\author{
Basile Lewandowski\thanks{Corresponding author. {\footnotesize \tt name.surname@unine.ch}} }
\author{
Lydia Y. Chen 
}
\affil{University of Neuchâtel}

\begin{document}
\maketitle

\begin{abstract}
\input{Sections/0_Abstract}
\end{abstract}

\section{Introduction}
\input{Sections/1_Intro}

\section{Related Studies}
\input{Sections/2_SOTA}

\section{Preliminaries}
\input{Sections/3_Preliminaries}

\section{\algo}
\input{Sections/4_Method}

\section{Experimental Results}
\input{Sections/5_Experiments}

\section{Conclusion}
\input{Sections/6_Conclusion}

\clearpage

{
    \small
    \bibliographystyle{ieeenat_fullname}
    \bibliography{references}
}

\clearpage
\setcounter{page}{1}
\maketitlesupplementary
\onecolumn
\input{Sections/A_Supp}

\end{document}

%% file: preamble.tex
\makeatletter
\robustify\@latex@warning@no@line
\makeatother
\usepackage{authblk}

\newcommand{\algo}{\textsf{MPQ-Diff}\xspace}

\newif\ifcomm

\usepackage[pagebackref,breaklinks,colorlinks,allcolors=cvprblue]{hyperref}
\usepackage[dvipsnames]{xcolor}
\definecolor{BlueIMT}{RGB}{0,42,72}
 \hypersetup{
    colorlinks=true,
    urlcolor=BlueIMT,
    linkcolor=MidnightBlue,
    citecolor=Green
 }
\usepackage{amsthm,amsmath,amssymb,amscd,mathrsfs}
\usepackage{txfonts,amsfonts,bbm}
 \usepackage{multirow}
 \usepackage{arydshln}

\usepackage[capitalize,noabbrev]{cleveref}

\usepackage{crossreftools}
\usepackage{algorithm}
\usepackage[noend]{algpseudocode}

%% file: Sections/0_Abstract.tex
Diffusion models (DMs) generate remarkable high quality images via the stochastic denoising process, which unfortunately incurs high sampling time. Post-quantizing the trained diffusion models in fixed bit-widths, e.g., 4 bits on weights and 8 bits on activation, is shown effective in accelerating sampling time while maintaining the image quality. Motivated by the observation that the cross-layer dependency of DMs vary across layers and sampling steps, we propose a \textbf{mixed precision quantization scheme}, \algo, which allocates different bit-width to the weights and activation of the layers.  We advocate to use the cross-layer correlation of a given layer, termed network orthogonality metric, as a proxy to measure the relative importance of a layer per sampling step.  We further adopt a uniform sampling scheme to avoid the excessive profiling overhead of estimating orthogonality across all time steps. We evaluate the proposed mixed-precision on LSUN and ImageNet, showing a significant improvement in FID from 65.73 to 15.39, and 
52.66 to 14.93, compared to their fixed precision quantization, respectively.

%% file: Sections/1_Intro.tex
Diffusion models (DMs) recently emerg as the state-of-the-art approach for image generation, surpassing generative adversarial networks (GANs) and variational autoencoders (VAEs) in both quality and diversity~\cite{DM_beat_GAN, HQ_Image_Syn}.
To synthesize images, DMs gradually add noise to an image and then learn to reverse this process, ultimately enabling the generation of high-quality images from pure noise~\cite{luo2022understandingdiffusionmodelsunified}. 
The key obstacle of deploying DMs arises from their computationally intensive iterative denoising process~\cite{DDPM, q_diff, PTQD} during the sampling. 
To address these challenges,  various approaches are explored to accelerate and compress DMs. Advanced sampling techniques are shown successfully to reduce the number of required denoising steps from thousands to mere dozens without significant quality loss~\cite{Implicit_DM, Improved_DM, song2020improvedtechniquestrainingscorebased, wang2023diffusiongantraininggansdiffusion}. However, each inference step still demands considerable computational resources \cite{NEURIPS2023_e4667dd0}. Moreover, as DMs grow in complexity and therefore in parameter count, their memory footprint expands to multiple gigabytes.
Model compression techniques, such as quantization, pruning, and distillation, are proposed to tackle these issues \cite{shang2023ptqdm, fang2023structuralpruningdiffusionmodels, Distillation_DM}.

Among these, quantizing the model parameters into lower bit representation not only reduces model size but also accelerates inference speed~\cite{EfficientDM, PTQD, so2023temporaldynamicquantizationdiffusion}. 
To the best of our knowledge, existing quantization schemes for DM focus on the fixed representation, where all layers of the model are quantized into the same bit-widths, typically 4 or 8 bits~\cite{q_diff, shang2023ptqdm, PTQD, EfficientDM}. For classification models, 
mixed-precision quantization assigns different bit-widths to different layers of the network, offering a more fine-grained trade-off between compression ratio, model size and model performance~\cite{dong2019hawqv2hessianawaretraceweighted,wang2019haqhardwareawareautomatedquantization,BRECQ}, and even outperforming fixed-precision quantization~\cite{OMPQ}. 
Despite its potential benefits, mixed-precision quantization for DMs remains largely unexplored. 

The central research challenge of mixed precision is how to allocate the available memory space to different layers of the networks based on their importance. Different from classification models, the application of mixed-precision quantization to DMs presents unique challenges due to the progressing nature of the sampling process. In contrast to standard neural networks, where the importance of each layer remains constant during inference~\cite{yang2020fracbitsmixedprecisionquantization, BRECQ, OMPQ}, the contribution of different layers in a DM varies across denoising time steps. This complicates the task of determining optimal bit-width allocations and requires a novel approach tailored to the specific characteristics of DMs.

\begin{figure*}[htp]
    \centering
    \begin{subfigure}{0.3\linewidth}
        \centering
        \includegraphics[width=\linewidth]{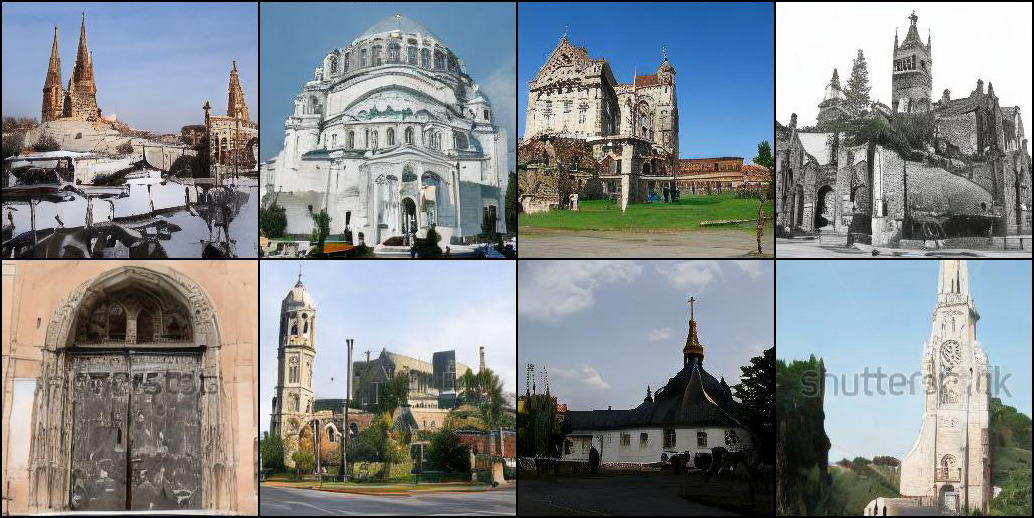}
        \caption{Full Precision }
    \end{subfigure}%
    \hfill
    \begin{subfigure}{0.3\linewidth}
        \centering
        \includegraphics[width=\linewidth]{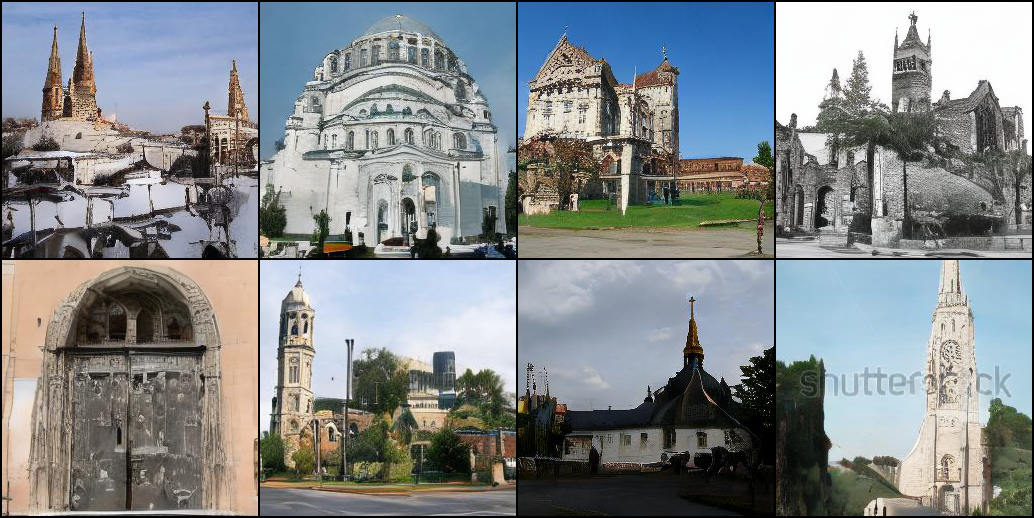}
        \caption{EfficientDM (Fixed quantization with 4 bits weights \& 8 bits activation). }
    \end{subfigure}%
    \hfill
    \begin{subfigure}{0.3\linewidth}
        \centering
        \includegraphics[width=\linewidth]{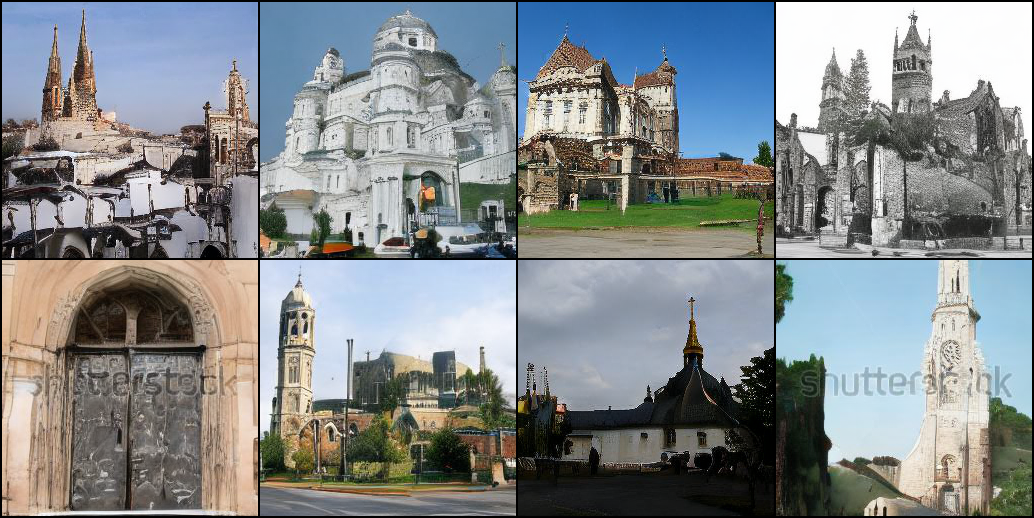}
        \caption{Mixed precision of \algo }
    \end{subfigure}
    \caption{Samples of generated images under different bit precision. 
    }
    \label{fig:generated_img}
\end{figure*}

In this paper, we fill the research gap by proposing the first mixed-precision quantization framework, termed \algo, specifically designed for diffusion models. Our approach builds upon the concept of ORrthogonality metric, termed ORM,~\cite{OMPQ}, which quantifies the correlation of any two layers of different dimensions. A layer that has higher aggregated ORM shows a higher correlation with other layers, indicating the higher relative importance compared to the layers with lower aggregated ORM.  We adapt this concept to the iterative structure of DMs, developing a timestep-aware bit-width allocation method that accounts for the varying importance of layers, approximated by their aggregated ORM, across the denoising steps. 
These metrics are then aggregated and used to solve a linear programming problem that returns bit-width allocation while integrating model size constraints. The proposed \algo is compatible to any fixed-width quantization techniques, demonstrated in our evaluation. 

The main contributions are as follows:

\begin{enumerate}
\item We identify the unique challenges of mixed-precision quantization for diffusion models, including the varying importance of layers across timesteps and the computational and memory cost of analyzing all timestep data.

\item We introduce a timestep-aware bit-width allocation strategy based on network orthogonality. This approach takes into account the dynamic nature of diffusion models by analyzing layer importance across all timesteps, resulting in more effective bit-width allocations for diffusion models.

\item We develop an efficient uniform sampling technique that enables comprehensive analysis of layer importance across all timesteps without incurring excessive computational and memory overhead.

\item We conduct extensive experiments on LSUN and ImageNet datasets, integrating our approach with existing fixed-quantization techniques. Our method outperforms fixed quantization, reducing FID by 2.85 and 2.57 points (on PTQD and EfficientDM respectively) on LDM-4 for ImageNet 256×256. With only ~10\% increase in model size, we dramatically improve FID from 65.73 to 15.39 (LSUN-Churches) and 52.66 to 14.93 (LSUN-Bedrooms).

\end{enumerate}

%% file: Sections/2_SOTA.tex
\textbf{Diffusion Model Quantization} While diffusion models can produce high-quality samples, their slow generation speed hinders large-scale applications. Quantization has emerged as a popular approach to reduce both memory footprint and inference time for diffusion models. These approaches can be broadly categorized into post-training quantization (PTQ) \cite{nagel2020downadaptiveroundingposttraining, BRECQ, wei2023qdroprandomlydroppingquantization, hubara2020improvingposttrainingneural} and quantization-aware training (QAT) \cite{   esser2020learnedstepsizequantization, gong2019differentiablesoftquantizationbridging, louizos2018relaxedquantizationdiscretizedneural, zhuang2017effectivelowbitwidthconvolutionalneural, nagel2022overcomingoscillationsquantizationawaretraining}.
PTQ is an offline quantization method that does not require access to the full training dataset, making it highly practical and easy to implement. Q-Diffusion was the first to achieve promising results on quantizing diffusion models with a PTQ approach \cite{q_diff}. Subsequent works such as PTQD, PTQ4DM, and EfficientDM have focused on improving results and obtaining even higher quality images with lower bit-width quantization \cite{PTQD,shang2023ptqdm,EfficientDM}.
In contrast, QAT adopts an online quantization strategy, utilizing the entire training dataset during the quantization process. While this can lead to superior accuracy, its adoption is limited to scenarios where retraining/fine-tuning is feasible.

\textbf{Mixed precision quantization} Quantization strategies can be further divided into fixed-bit quantization and mixed precision quantization \cite{Wang_2019_CVPR}. Fixed quantization allocates the same number of bits to each layer of the model, while mixed precision quantization allows for different bit-widths across layers \cite{Wang_2019_CVPR}.
Mixed precision quantization aims to achieve a better trade-off between compression ratio and accuracy by assigning different bit-widths to different network layers \cite{BRECQ,OMPQ}. This approach has been well-researched for convolutional neural networks, with BRECQ introducing the idea for ResNet architectures \cite{BRECQ}. OMPQ built upon this work to develop an even more effective method based on network orthogonality to assign bit-widths to a model's layers \cite{OMPQ}.
However, the application of mixed precision quantization to diffusion models remains largely unexplored. The iterative nature of diffusion models presents unique challenges that require adapting existing mixed precision techniques.

%% file: Sections/3_Preliminaries.tex
\subsection{Diffusion models}
Diffusion models \cite{Implicit_DM, DDPM} are a class of generative models that iteratively introduce noise to real data $\mathbf{x}_0$ through a forward process and generate high-quality samples via a reverse denoising process. Unlike other image generation models such as VAEs and GANs, Diffusion Models generate images through an iterative, time-step specific process. The same U-Net architecture is employed across all generational steps, reusing parameters throughout the denoising process \cite{luo2022understandingdiffusionmodelsunified}. Let us consider an image $\mathbf{x}$ (or its latent representation) that is increasingly noised over time following a Gaussian distribution $\mathcal{N}$. Generally, the forward process is a Markov chain, which can be formulated as: 
\begin{align}
q(\mathbf{x}_t|\mathbf{x}_{t-1}) = \mathcal{N}(\mathbf{x}_t; \sqrt{\alpha_t}\mathbf{x}_{t-1}, \beta_t \mathbf{I}), \nonumber \\ 
q(\mathbf{x}_{1:T}|\mathbf{x}_0) = \prod_{t=1}^{T} q(\mathbf{x}_t|\mathbf{x}_{t-1}), 
\end{align}
where $\alpha_t, \beta_t$ are hyperparameters and $\beta_t = 1 - \alpha_t$.

The reverse process is intractable, since we cannot directly estimate the real distribution of $q(\mathbf{x}_{t-1}|\mathbf{x}_t)$. Diffusion models hence approximate the distribution via variational inference by learning a Gaussian distribution: 
\begin{equation}
p (\mathbf{x}_{t-1}|\mathbf{x}_t) = \mathcal{N}(\mathbf{x}_{t-1}; \mu(\mathbf{x}_t, t), \Sigma(\mathbf{x}_t, t)), 
\end{equation}
where $\mu$ and $\Sigma$ are two neural networks. The mean $\mu(\mathbf{x}_t, t)$ can be reparameterized by a noise prediction network $\epsilon(\mathbf{x}_t, t)$ as follows:
\begin{equation}
\mu(\mathbf{x}_t, t) = \frac{1}{\sqrt{\alpha_t}} \left( \mathbf{x}_t - \frac{\beta_t}{\sqrt{1-\bar{\alpha}_t}} \epsilon(\mathbf{x}_t, t) \right),
\end{equation}
where $\bar{\alpha}_t = \prod_{s=1}^{t} \alpha_s$.

The variance $\Sigma(\mathbf{x}_t, t)$ can either be reparameterized or fixed to a constant schedule $\sigma_t$. When using a constant schedule, the sampling of $\mathbf{x}_{t-1}$ can be formulated as:
\begin{equation}
\mathbf{x}_{t-1} = \frac{1}{\sqrt{\alpha_t}} \left( \mathbf{x}_t - \frac{\beta_t}{\sqrt{1-\bar{\alpha}_t}} \epsilon(\mathbf{x}_t, t) \right) + \sigma_t \mathbf{z}, \quad \text{where} \, \mathbf{z} \sim \mathcal{N}(\mathbf{0}, \mathbf{I}).
\end{equation}

\subsection{Quantization}
Quantization is a technique used to map high-precision model parameters and activations, typically floating-point numbers, to lower-precision integer values. 
Given a floating-point vector \(x\), quantization maps its values to discrete points (or grids). The quantization function is generally designed to minimize the quantization error, which is the difference between the original value \(x\) and its quantized counterpart \(\hat{x}\). This can be expressed as:

\[
\hat{x} = Q_U(x, s) = \text{clip}\left(\left\lfloor \frac{x}{s} \right\rfloor, l, u \right) \cdot s,
\]

where \(\lfloor \cdot \rfloor\) is the round operation, \(s\) is the step size (or scale), and \(l\) and \(u\) are the lower and upper bounds of the quantization thresholds.

The default quantization method is uniform symmetric quantization, where these grids are evenly spaced and symmetrically distributed.
Uniform quantization can be described as:

\[
\hat{x} = \Delta \cdot \left( \text{clip}\left(\left\lfloor \frac{x}{\Delta} \right\rfloor + Z, 0, 2^b - 1 \right) - Z \right),
\]

where \(\Delta = \frac{\max(x) - \min(x)}{2^b - 1}\), and \(Z = - \left\lfloor \frac{\min(x)}{\Delta} \right\rfloor\). $b$ denotes the bit-width of the quantized variable, i.e. it can have $2^b$ different values. Mixed precision quantization is about finding a balance between small bit-width with low precision and higher bit-with implying larger memory footprint.

\subsection{Network Orthogonality}\label{pt:orthogonality}
Neural Networks can be decomposed into a set of layers or functions $\mathcal{F}$. i.e., $\mathcal{F} = \{f_1, \ldots, f_i, \ldots, f_L\}$ where $f_i$ represents the transformation from the input of the model to the i-th layer.
According to \cite{arfken2011mathematical}, if $\langle f_i, f_j\rangle = 0$, then $f_i$ and $f_j$ are weighted orthogonal.
Previous works have argued the quantization error can reach 0 if $\mathcal{F}$ is an orthogonal basis function set \cite{OMPQ}. Furthermore, it claims strong orthogonality between the basis functions yields a stronger representation capability for the corresponding model.
To allow for simple orthogonality computation, the authors proposed a sampling strategy to approximate orthogonality with a so-called ORthogonality Metric (ORM) defined as:
\begin{equation}
\label{eq:orm_function}
\operatorname{ORM}(X, f_i, f_j) = \frac{\|f_j(X)^\top f_i(X)\|_F^2}{\|f_i(X)^\top f_i(X)\|_F\|f_j(X)^\top f_j(X)\|_F}
\end{equation}

where ORM $\in$ [0, 1] and  $\| \cdot \|_F$ is the Frobenius norm.  $f_i$ and $f_j$ are orthogonal when ORM = 0 and dependent when ORM = 1. In practice, we compute ORM as a higher level dot product that account not only for alignment but for relative corelation across the layers.

Therefore, ORM is negatively correlated to orthogonality. Notice the metric we obtain is the same as the linear Centered Kernel Alignement (CKA) which was introduced to explore similarity between different hidden layers \cite{CKA}. 
Our goal is here to maximize the representation capability of the quantized neural network by assigning a larger bit-width to the layers with stronger orthogonality against all other layers.

There are several key advantages of using this metric to construct a quantization algorithm. First, we do not require running samples through the model multiple time, as opposed to dynamic quantization. One set of activation values is sufficient to decide the bit-width allocation and we do not require further training for to setup the quantization. Moreover, orthogonality gives an insight of the correlation between the layers. This feature is preeminent as layers that need a higher precision are not only the ones more prone to errors but also the ones more deeply linked with the rest of the network. This matter is utmost in diffusion models where the correlation between layers is not straightforward because of their architecture (typically U-Net \cite{unet}) that often includes different skip connections. 

%% file: Sections/4_Method.tex
\begin{figure*}
\hypertarget{fig:overview}{}
\centering
\includegraphics[trim={18pt 40pt 50pt 86pt}, clip, width=\linewidth]{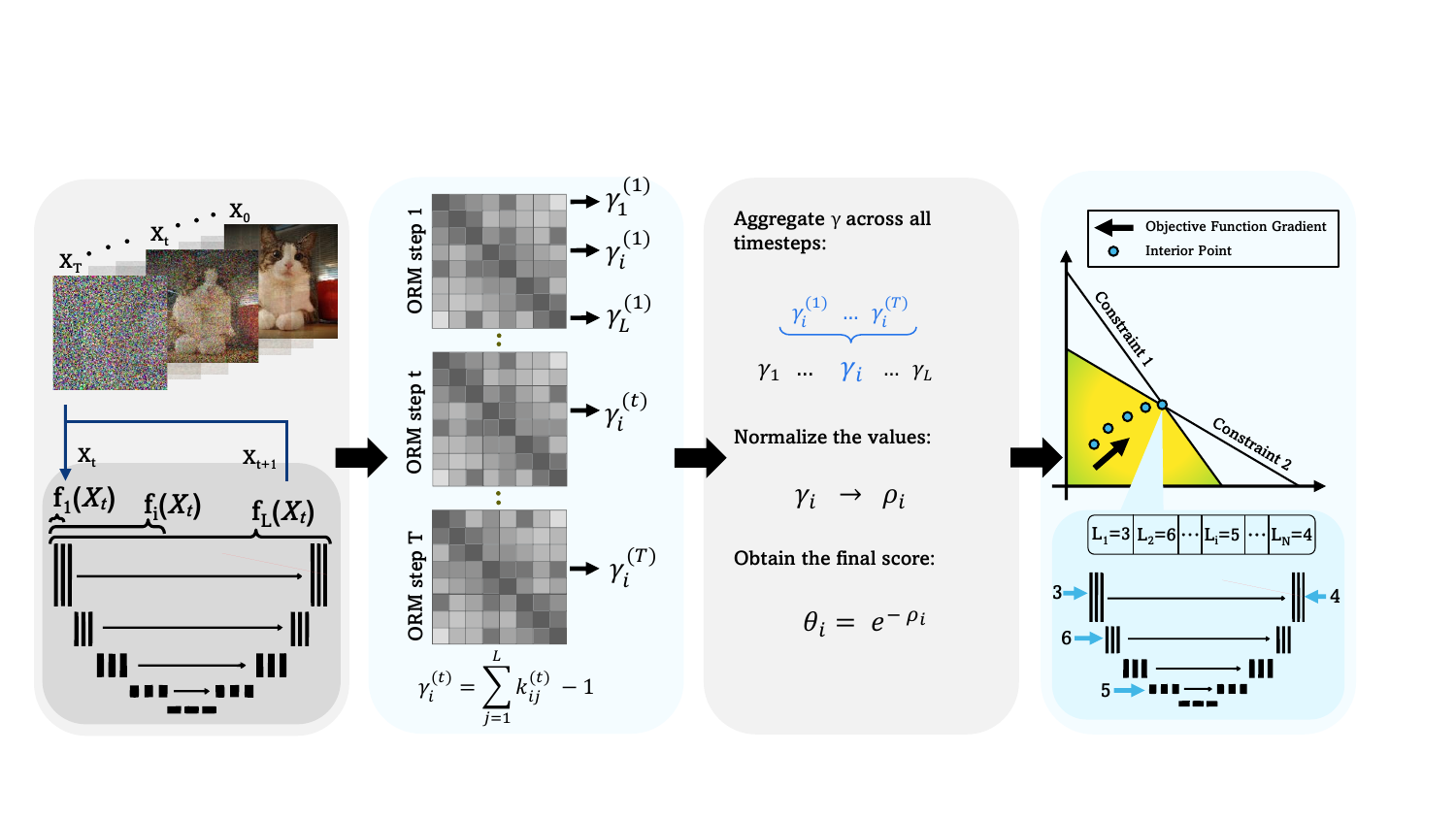}

\vspace{0.5em}

\begin{minipage}[t]{0.23\textwidth}
  \centering
  \textbf{a) Deconstruction of DM}
\end{minipage}%
\hfill
\begin{minipage}[t]{0.23\textwidth}
  \centering
  \textbf{b) ORMs Calculation}
\end{minipage}%
\hfill
\begin{minipage}[t]{0.23\textwidth}
  \centering
  \textbf{c) Timestep Aggregation}
\end{minipage}%
\hfill
\begin{minipage}[t]{0.23\textwidth}
  \centering
  \textbf{d) LPP Evaluation}
\end{minipage}
\caption{Overview of the \algo workflow. a) Deconstruct the DM into a set of functions $\mathcal{F}$, which are used across all $T$ generation timesteps. b) The ORM matrices for every sampled timestep is calculated from $\mathcal{F}$.
c) Aggregation of all ORM matrices to obtain overall function importance across timesteps. d) LPP constructed by the importance factor $\theta$ to derive bit configuration.}
\label{fig:overview}
\end{figure*}

We  now introduce \algo, which leverages the aggregated layer correlation to quantize layers of diffusion models in different bit-width. \algo takes into account both network orthogonality and the iterative nature of the diffusion process to determine an efficient memory allocation under mixed precision quantization. The different steps of this process are illustrated in \autoref{fig:overview}.

As the ORM has the advantage of capturing correlation between any pair of layers of different dimension, it is well suited for underlying the correlation of UNets - the typical network architecture for diffusion models. Hence, we propose to use the ORM as the basic building block to measure the importance of different layers of DMs. 
The structure of the diffusion process however requires that we take into account not only the different layers of the network but also the different timesteps of the process. We therefore introduce in the following the challenge of aggregating ORM values across time and we will then detail how we use them to determine the bit-width allocation.

\subsection{Time-dependent ORM } 
\begin{figure}[h]
\centering
\includegraphics[width=\linewidth]{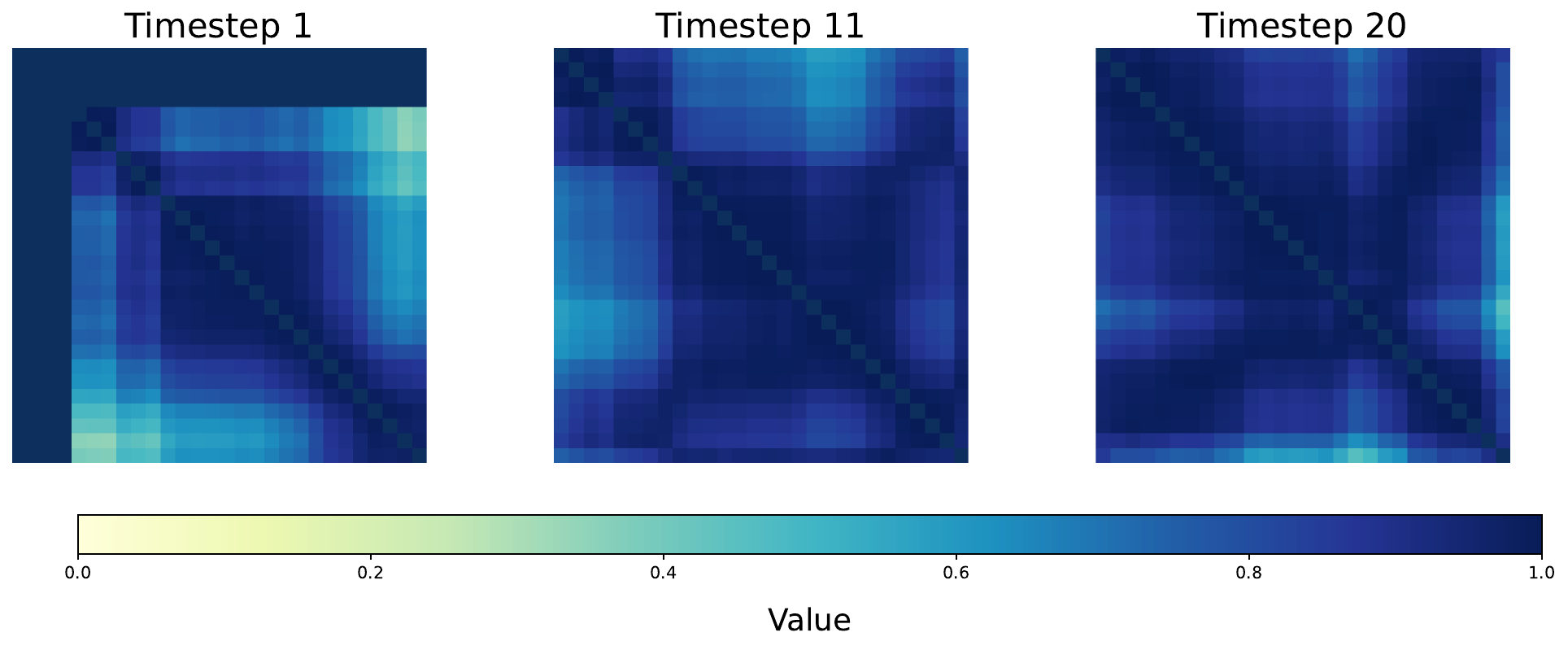}
\caption{Orthogonality Matrices across timesteps, for LDM-4 on Imagenet (steps = 20,
eta = 0.0,
scale = 3.0)}
\label{fig:ORM_variability}
\end{figure}

The success of mixed-precision quantization relies on the ability to identify important layers within a given model architecture and allocate bit-widths accordingly, providing more precision to performance-critical layers. Existing methods for traditional single-timestep models, such as BRECQ \cite{BRECQ} and OMPQ \cite{OMPQ}, utilize the outputs of individual layers or blocks to assign importance scores, which then inform bit-width allocation. However, diffusion models exhibit varying activation ranges across their time-steps due to the differing inputs at each step of the denoising process. Consequently, different layers have different importance at different steps of the denoising process. This phenomenon is illustrated in \autoref{fig:ORM_variability}, which shows how the ORM values which we use for bit-width allocation, vary significantly across timesteps for a diffusion model.

This observation aligns with previous research that attributes specific functions to different layers within the U-Net architecture of diffusion models \cite{Improved_DM}. In particular, the behavior of input layers differs markedly from that of middle and output layers. As a result, naively selecting a single timestep to determine bit-width allocation based on layer outputs is insufficient in our case.

\begin{figure}
\centering
\includegraphics[width=\linewidth]{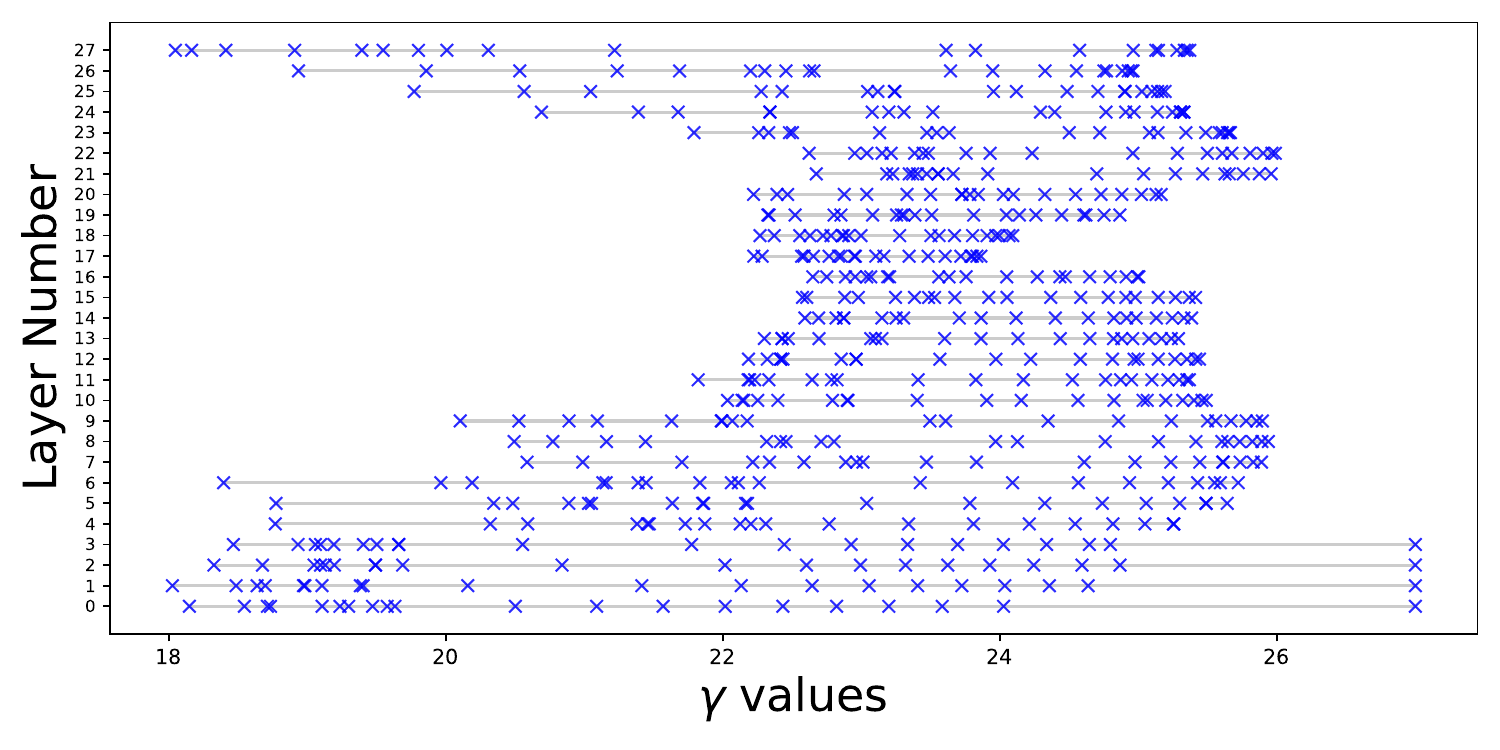}
\caption{ Different $\gamma$ coefficients (one for each timestep) per layer on LDM-4 Imagenet (steps = 20 eta = 0.0 scale = 3.0). It appears the inner layers show smaller variance over time. }
\label{fig:layer_importance_across}
\end{figure}

Therefore, we propose a new method to aggregate the importance of a layer across all timesteps, while balancing the significance of outlier values and global importance. 
This method offers several key advantages. Firstly, it balances local and global importance by considering both critical timesteps where a layer may have unusually high importance and its overall contribution across the generation process. The exponential weighting gives more emphasis to timesteps with lower orthogonality scores (indicating higher importance) while still accounting for all timesteps.    
By capturing the dynamic nature of layer importance throughout the entire process, we obtain a more comprehensive assessment of each layer's role.  \autoref{fig:layer_importance_across} shows the varying importance of layers throughout all 20 timesteps for the LDM-4 model on Imagenet 256 × 256. It appears that all layers have quite a large range of $\gamma$ values depending on the timestep selected to measure it, raising the question of their aggregation.

Our proposed method aims to improve upon simple averaging across all timesteps. While averaging treats all timesteps equally, potentially diluting the impact of critical generation iterations, this weighted approach preserves the significance of key timesteps while still considering the overall trend. This is particularly important in diffusion models, where certain layers may play crucial roles at specific points in the generation process~\cite{PTQD}.
In summary, our time-aware layer importance scoring algorithm provides a more accurate and comprehensive assessment of layer roles in diffusion models. By carefully balancing local criticality with global significance, we obtain a better estimation of each layer's contribution to the image generation process across timesteps, enabling more informed optimization strategies.
\label{pt:timesteps}
\begin{algorithm}

\caption{Bit-Width Allocation}
\label{alg:complete_alg}
\textbf{Require:} Pretrained full precision diffusion model\\
\textbf{Require:} Normalization algorithm \verb|ZScore| \\
\textbf{Require:} Generation of $n$ samples $x_1, \ldots, x_n$ which are used to construct the ORM matrices.\\
\textbf{Require:} Calculation of ORM matrices $\{K^{(1)}, \ldots, K^{(T)}\}$
\begin{algorithmic}
\For{layers $1, \ldots, i, \ldots, L$}:
    \State $\gamma_{i} \gets \{\sum_{j=1}^{L} k^{(t)}_{ij} - 1\}_{1<t<T}$
    \State $\Tilde{\gamma}_{i} \gets \verb|ZScore| (\gamma_i)$ 
    \State $w_{i} \gets e^{-\Tilde{\gamma}_{i} }$ 
    \State $\rho_i \gets \frac{\gamma^\top w}{\|w\|}$
    \State $\theta_i \gets e^{- \rho_i}$ 
\EndFor
\State Construct and solve the Linear Programming Problem with $\{\theta_i\}_{i=1}^L$ as input (see \autoref{eq:constraints})
\end{algorithmic}
\textbf{Returns:} per-layer bit-width allocation

\end{algorithm}
\subsection{Mixed precision quantization }
The simplest use of quantization methods is to apply a given bit-width to all layers in a model. However, this approach overlooks the different sensitivities of each layer, with some of them being more susceptible to quantization error propagation than others. Mixed Precision Quantization addresses this issue by assigning different bit widths to the layers under consideration. This allows benefiting from low-bit quantization performance where it's possible while preserving precision for layers that require it.

In order to allocate bits to the different layers  $\mathcal{F} = \{f_1,  \ldots, f_L\}$ of a diffusion model, we first decompose the pre-trained diffusion model in $\{f_i\}_{i=1}^L$ layers and compute their respective importance. We generate a batch of images using $T$ denoising steps and consider the output of every single layer (see \hyperlink{fig:overview}{Figure~\ref{fig:overview}a}). With these values, we can construct a matrix of all ORM coefficients according to \autoref{eq:orm_function}. Considering a timestep $t$ and an input $X$, we define the ORM matrix $K^{(t)}$ such as $k^{(t)}_{i,j} := \operatorname{ORM}(X^{(t)}, f_i, f_j)$. $K$ is then a symmetric matrix with diagonals values equal to one where a given row represents the orthogonality metrics of a layer with regard to every other ones. 

Therefore we construct $T$ ORM matrices (one for each timestep), and add up the non-diagonal elements of each row to account for the cumulated orthogonality of the different layers. 
\begin{equation}
\gamma_i^{(t)}=\sum_{{1\leq j \leq L}\atop{j \neq i}} k^{(t)}_{i,j}    
\end{equation}

We then compute their normalized counterpart $\rho$ to account for the relative importance of the layer. Finally, we apply an exponential decrease to highlight the effect of outliers and obtain the $\theta$ coefficients. The whole process is detailed in \autoref{alg:complete_alg}.
The final values are used to construct and solve the Linear Programming Problem (LPP) proposed in earlier works \cite{OMPQ}. 
The $\{\theta_i\}_{i=1}^L$ are used as the importance factor for the layers of the DM and the LPP is defined as follows:
\begin{align}
\text{Objective: }& \max_{\mathbf{b}} \sum_{i=1}^{L} \left(\frac{b_i}{L-i+1} \sum_{j=i}^{L} \theta_j\right) \label{eq:objective_function} \\
\text{Constraints: }& \sum_{i} \operatorname{size}(M_i, b_i) \leq \mathcal{T} \label{eq:constraints}
\end{align}

Maximizing aggregated orthogonality  \eqref{eq:objective_function} is taken as the objective function, which is employed to integrate the model size constraints \eqref{eq:constraints}.
The objective is to find the optimal bit-width distribution $\mathbf{b}^* \in B^L$ that maximizes the scale of the most important layers, with $B$ being the different bit-widths considered (typically three to eight bits).
The constraint 
is expressed as keeping the cumulated memory size of every layer  (i.e. $\operatorname{size}(M_i, b_i)$) below a specified target $\mathcal{T}$. 
Maximizing the objective function means assigning the larger bit-widths to more independent layer, which, as was outlined in \autoref{pt:orthogonality}, implicitly maximizes the model’s representation capability. The LPP is a classic optimization problem and can easily be solved using a negligible amount of computing resources \cite{slsqp}.

\begin{figure}
\centering
\includegraphics[width=\linewidth]{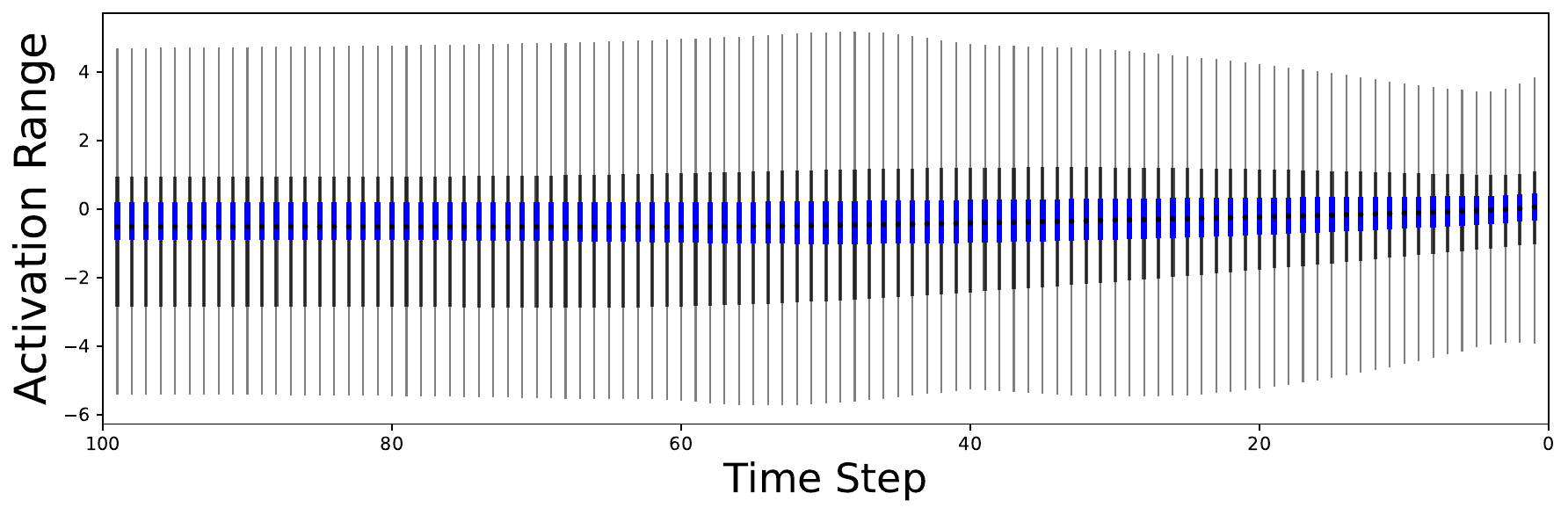}
\caption{Activation ranges of $x_t$ across all 100 time steps of LDM-4 model on Imagenet 256 × 256. The blue regions represent the inter-quartile range (first to third quartile) of activation values, while the gray regions extend from the 5th to 95th percentiles. 
}
\label{fig:activation_ranges}
\end{figure}

\subsection{Estimating ORM by Sampling}

\begin{figure}[h]
\centering
\includegraphics[width=\linewidth]{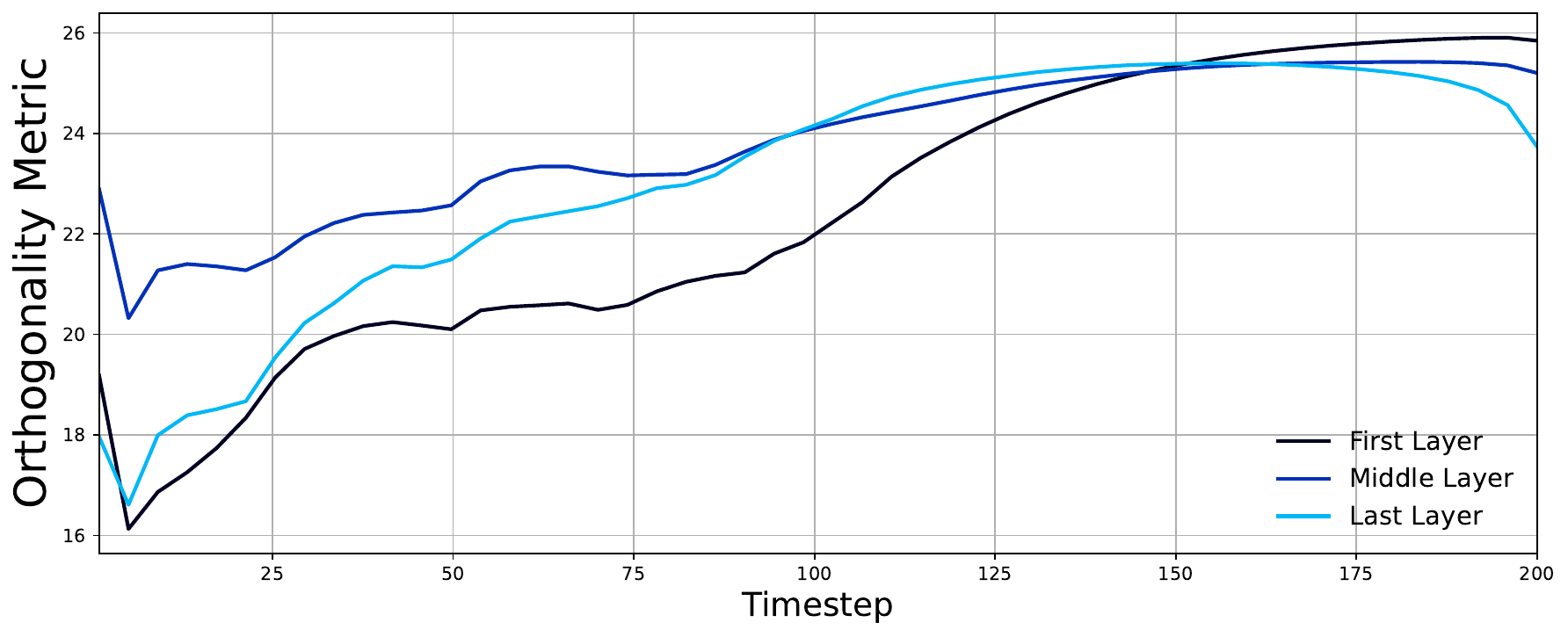}
\caption{$\theta_i$ across 200 generation timesteps on LDM-4 Imagenet 256 × 256 (steps = 200 eta = 0.0 scale = 3.0).}
\label{fig:orthogonality_across}
\end{figure}

 To comprehensively assess the importance of each layer across all timesteps and determine an optimal bit-width allocation, one would ideally collect output data from every layer at each timestep of the diffusion process. However, this approach proves to be extremely memory-intensive and computationally demanding. Collecting complete data might be feasible for models using a few denoising steps (e.g. 20), but the memory requirements become impractical for models with a higher number of generation steps. For instance, saving data for 200 generation steps from LDM-4 on ImageNet 256 × 256 would require approximately 150 gigabytes of memory. Additionally, calculating any metric across such a number of timesteps slows down the process of identifying an appropriate bit-width allocation significantly. A key insight that can help address the memory and computational challenges is the observation that activations of the same layer tend to be quite similar in neighboring timesteps. This property of diffusion models has been noted in previous works \cite{q_diff} and further corroborated by subsequent studies \cite{PTQD}. \autoref{fig:activation_ranges} shows this phenomenon on LDM-4 with Imagenet. It suggests that the importance of a layer, while varying across the entire denoising process, changes gradually rather than abruptly between adjacent steps. \autoref{fig:orthogonality_across} further supports this claim, highlighting the similarity of orthogonality values for layers across neighboring timesteps.

This observation suggests that while comprehensive data collection across all timesteps would be ideal, it is not be necessary to achieve a near-optimal bit-width allocation. We subsequently introduce an efficient sampling strategy to optimize our layer importance assessment algorithm for diffusion models with numerous timesteps. This approach addresses the computational and memory challenges associated with models requiring many denoising iterations, while maintaining the integrity of our analysis. For diffusion models with more than 50 timesteps, we implement a uniform sampling method to select a subset of timesteps for analysis. Our sampling approach offers several key benefits:

\begin{enumerate}[itemsep=-2pt, topsep=0pt, partopsep=0pt]
\item Memory Efficiency: By reducing the number of timesteps analyzed, we significantly decrease the memory requirements for storing intermediate outputs. This is crucial for scaling our method to more complex models and longer generation sequences.
\item Computational Speed: Analyzing fewer timesteps naturally leads to faster computation times. This efficiency gain allows for more rapid experimentation and search for different bit-width allocations.
\item Generalizability: Our sampling strategy makes our method adaptable to a wide range of diffusion models, including those that require hundreds or thousands of denoising steps. 
\item Preserved Accuracy: Despite reducing the number of analyzed timesteps, our method maintains a high level of accuracy in assessing layer importance. The uniform sampling ensures that we capture the overall trends and critical points in the generation process.
\end{enumerate}

This sampling method effectively balances the trade-off between computational efficiency and analytical depth. By capturing a diverse set of timesteps across the generation process, we ensure that our analysis remains representative of the model's behaviour throughout denoising.

%% file: Sections/5_Experiments.tex

\textbf{Models and Metrics.}
To verify the effectiveness of the proposed method, we conduct extensive image synthesis using Latent Diffusion Models (LDM)~\cite{5206848} on three standard benchmarks: ImageNet~\cite{HQ_Image_Syn}, LSUN-Bedrooms, and LSUN-Churches~\cite{yu2016lsunconstructionlargescaleimage}, using images with a resolution of 256 × 256. All experimental configurations, including the number of steps, variance schedule (denoted by eta in the following), and classifier-free guidance scale, follow the parameterization and implementation of the respective papers. The performance of diffusion models is evaluated with Fréchet Inception Distance (FID) and Sliding Fréchet Inception Distance (sFID) \cite{heusel2018ganstrainedtimescaleupdate, salimans2016improvedtechniquestraininggans}. Results are obtained by
sampling 50'000 images and evaluated with ADM’s TensorFlow evaluation suite \cite{DM_beat_GAN}. All experiments are run on a Nvidia RTX3090 GPU and implemented with the PyTorch framework.

\textbf{Quantization Implementation.}
To verify the adaptability of our method, we implement mixed precision quantization on two fixed-precision quantization methods proposed in the literature. PTQD~\cite{PTQD} and EfficientDM~\cite{EfficientDM} are effective in fixed-precision quantization and can be adapted to for mixed-precision quantization (allocating differing bit-width to different layers). 
The input embedding and output layers of the models employ a fixed W8A8\footnote{8 bits for weights and 8 bits for activation} quantization, whereas other convolution and fully-connected layers are quantized to the target bit-width.
For EfficientDM experiments, the rank of the adapter is set to 32, LoRA weights and quantization parameters are finetuned for 12.8K iterations with a batchsize of 4, as in the original work~\cite{EfficientDM}.

\subsection{Class-conditional Generation}

We begin our evaluation by performing conditional generation on the ImageNet dataset and implement our method using the baselines (PTQD and EfficientDM) as quantization algorithm. The results are
presented in \autoref{tab:result_perf}. 250 Mbs is chosen as the model size, as both PTQD and EfficientDM with quantization to 4 bits lead to this model size for LDM-4 on Imagenet. Therefore, we can compare the performance of mixed precision quantization to single precision quantization on a fixed model size that is achievable under the most popular quantization setting.
Our proposed method demonstrates superior performance compared to single precision quantization on the FID results. Notably, when implementing our approach with PTQD we achieve a FID decrease of 2.85 using the same model size and a FID decrease of 2.54 when implementing it on EfficientDM.
The results on sFID do not show a similar performance here, but the values of the quantizeed algorithm are already better or close to the ones from the full precision models, leaving small room for improvement.

\subsection{Unconditional Generation}

To further validate our method, We evaluate it on two standard benchmarks of unconditional generation: LSUN-Churches and LSUN-Bedrooms.
Here, the respective model size was chosen as the size achieved by quantization to 4 bits under PTQD and EfficientDM quantization. We show that our proposed aggregation method also outperforms single precision quantization in the uncontidional setting under both 4x and 8x downsampling factor in the latent space (LSUN-Bedrooms and LSUN-Churches respectively). For experiments implemented on EfficientDM, we implement 100 denoising steps and for experiments on PTQD 200 steps, to reflect the original  implementations. Our unconditional experiments use a sampling strategy of 50 steps to inform bit-width allocation (i.e. the aggregation uniformly samples 50 of the 100/200 timesteps to inform the bit-width allocation, as outlined in the methodology section). \autoref{tab:result_perf} shows that the proposed method achieves superior performance compared to fixed size quantization. For LSUN-Bedrooms we outperform single precision quantization results on FID by 0.31 and 1.62 when implemented on PTQD and EfficientDM respectively. For LSUN-Churches we improve the FID by 1.14 and 0.34 when implemented on PTQD and EfficientDM respectively.

\begin{table*}
 \caption{Performance comparisons of mixed precision quantization across various settings. For our method, we display in parenthesis the quantization algorithm used. Results of the full precision baseline are displayed in \textit{italic} and the best value for each experiment aside from them is displayed in \textbf{bold}. }
  \centering
  \begin{tabular}{l l l c c c c c}
    \toprule
    Dataset & Setup & Method & Model Size (Mb) & Bitwidth (W/A) & FID$\downarrow$ & sFID$\downarrow$ \\
    \cmidrule(lr){1-8}
    \multirow{5}{*}{\parbox{2cm}{ImageNet\\256×256}} 
        & \multirow{5}{*}{\parbox{1.2cm}{ LDM-4 \footnotesize{\\ steps = 20\\ eta = 0.0\\scale = 3.0}}} 
        & \textit{Full Precision} & \textit{1603} &\textit{ 32/32} & \textit{11.25} & \textit{7.70 }\\
    &  & PTQD & 250 & 4/8 & 9.83 & \textbf{6.40} \\
    &  & EfficientDM & 250 & 4/8 & 8.64 & 7.82 \\
    &  & Ours (PTQD) & 250 & MP/8 & 6.98 & 6.90 \\
    &  & Ours (EfficientDM) & 250 & MP/8 & \textbf{6.07} & 11.46 \\
    \cmidrule(lr){1-8}
    \multirow{5}{*}{\parbox{2.5cm}{LSUN-Bedrooms\\256×256}} 
        & \multirow{5}{*}{\parbox{1.3cm}{ LDM-4 \footnotesize{\\ steps = 200\\ eta = 1.0}}} 
        & \textit{Full Precision} & \textit{1096.2} & \textit{32/32} & \textit{3.00} & \textit{7.13} \\
    &  & PTQD & 194.26 & 4/8 & 6.34 & 15.78 \\
    &  & EfficientDM & 194.27 & 4/8 & 9.54 & 15.31 \\
    &  & Ours (PTQD) & 194.24 & MP/8 & \textbf{6.03} & 13.95 \\
    &  & Ours (EfficientDM) & 194.25 & MP/8 & 7.92 & \textbf{13.94} \\
    \cmidrule(lr){1-8}
    \multirow{5}{*}{\parbox{2.5cm}{LSUN-Churches 256×256}} 
        & \multirow{5}{*}{\parbox{1.3cm}{ LDM-8 \footnotesize{\\ steps = 200\\ eta = 0.0}}} 
        & \textit{Full Precision} & \textit{1179.9} & \textit{32/32} & \textit{6.30} & \textit{18.24} \\
    &  & PTQD & 180.2 & 4/8 & 8.38 & 18.92 \\
    &  & EfficientDM & 181.1 & 4/8 & 10.09 & 37.23 \\
    &  & Ours (PTQD) & 181.19 & MP/8 & \textbf{7.24} & \textbf{15.33} \\
    &  & Ours (EfficientDM) & 181.23 & MP/8 & 9.75 & 33.81 \\
    \bottomrule
  \end{tabular}
  \label{tab:result_perf}
\end{table*}

\subsection{Ablation Study}
To assess the efficacy of the proposed sampling method, we conduct a study on both the ImageNet and LSUN-Churches datasets using the LDM-4 model with a DDIM sampler, as presented in \autoref{tab:uniform_sampling}. We compute baseline $\theta$ values for bit-width allocation using all 200 timesteps, then calculate theta values by sampling 1/2, 1/4, 1/8, and 1/20 of the timesteps to find an optimal balance between accuracy and efficiency.
Our results show that sampling 1/4 of the timesteps offers an excellent tradeoff, achieving theta values close to our baseline while reducing memory and computational requirements by 75\%.
%

\begin{figure}
\centering
\includegraphics[width=\linewidth]{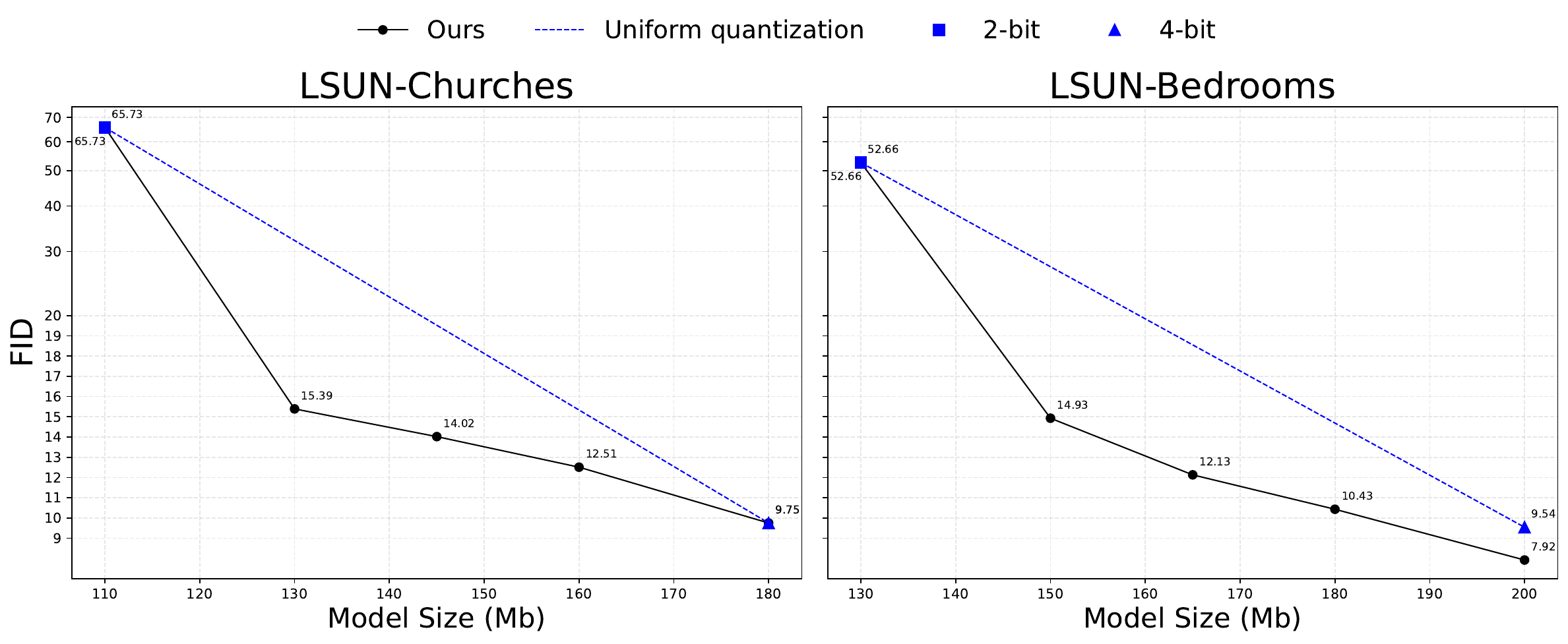}
\caption{FID for model sizes on LDM-4 LSUN-Churches, LSUN-Bedrooms (steps = 200 eta = 0.0 scale = 3.0)}
\label{fig:differing_size}
\end{figure}

We also compare our method with single precision quantization under different model sizes, with results reported in \autoref{fig:differing_size}. The graphs demonstrate that mixed precision quantization offers much more flexibility than single precision, allowing for model sizes that achieve the desired trade-off between performance and memory requirements.
A key observation from these results is how performance can improve with only a slight increase in model size, particularly when moving beyond fixed 2-bit quantization. Here, we observe a remarkable improvement in image quality with only a modest increase in model size. Specifically, the fixed 2-bit quantization for LSUN-Churches at 109 Mb yields a poor FID of 65.73. However, our mixed precision approach achieves a substantially better FID of 15.39 with only a slight increase in model size to 127 Mb. This represents a 76.6\% reduction in FID score for just a 16.5\% increase in model size.
This improvement highlights a critical inflection point where a small additional memory allocation can yield substantial gains in performance. Our mixed precision approach, by allowing more nuanced bit allocation, can thus preserve this critical information at minimal additional memory cost.
Furthermore, we can see that our method's performance at 127 Mb (FID 15.39) is already approaching that of 4-bit fixed quantization (FID 9.75 at 180 Mb), while using 29.4\% less memory. This indicates that performances close to 4-bit quantization can be achieved with significantly lower memory requirements using our mixed precision approach.

\begin{table}[h]

 \caption{Evaluation of Timesteps Sampling. With 1/4 sampling, the biggest change is only 3.23\% for LSUN-Churches and 3.03\% for ImageNet. Sampling 1/8 or fewer timesteps leads to larger deviations, potentially affecting bit-width allocation by the LPP. 
 }
 \resizebox{\columnwidth}{!}{%
  \centering
    \small 
  \begin{tabular}{c llll}
    \toprule
    Setup & Sampling & timesteps    &  MAPC    & Biggest change (\%)  \\
    \cmidrule{1-5}
    \multirow{5}{*}{\parbox{2.5cm}{\centering LSUN-Churches\\256 × 256 LDM-8 }}
    &  All timesteps  & 200 & --  &    -- \\
    & 1/2  & 100 & 0.42  &    1.20 $\pm$    0.21 \\
    & 1/4   & 50 & 1.12 & 3.23 $\pm$  0.56    \\
    & 1/8   & 25   & 2.22      & 6.21  $\pm$  1.08 \\
    & 1/20   & 10    & 4.10       & 10.21 $\pm$  1.85\\
    \midrule
    
    \multirow{5}{*}{\parbox{2.5 cm}{\centering ImageNet (Cond.)\\ LDM-4 256 × 256}}
    &   All timesteps & 200 & --  &       -- \\
    &  1/2  & 100 & 0.61  &  1.37  $\pm$  0.23  \\
    &  1/4   & 50 &  1.11 &  3.03 $\pm$ 0.51    \\
    &  1/8   & 25   &   2.43     &  7.19 $\pm$ 1.13  \\
    &  1/20   & 10    &    4.56   &  11.21 $\pm$  1.90 \\
    \bottomrule
  \end{tabular}
    \raggedbottom
  }
  \raggedright
  \footnotesize{MAPC = Mean Absolute Percentage Change}
  \label{tab:uniform_sampling}
  
\end{table}

%% file: Sections/6_Conclusion.tex
In this paper, we introduce mixed bit-width allocation strategy for diffusion models to account for the varying importance of layers across the entire denoising trajectory. To ensure the scalability on the large DMs,  we develop an efficient uniform sampling method to collect representative data across all timesteps without incurring prohibitive memory and computational costs. 
Through extensive experiments on both conditional and unconditional image generation tasks, \algo shows consistent improvements over fiexd-precision quantization methods across different model architectures and datasets. Specifically, the \algo reduces FID by 2.85 and 2.57 points on LDM-4 for ImageNet 256 × 256 compared to PTQD and EfficientDM quantization, respectively. Moreover, our results also show that a modest increase in model size can lead to dramatic improvements in generation quality. 
Our results highlight the effectiveness of mixed-precision, allowing for fine-grained control over the model size and performance trade-off.



%% file: Sections/A_Supp.tex

\begin{wrapfigure}{l}{0.5 \textwidth}
    \includegraphics[width=\linewidth]{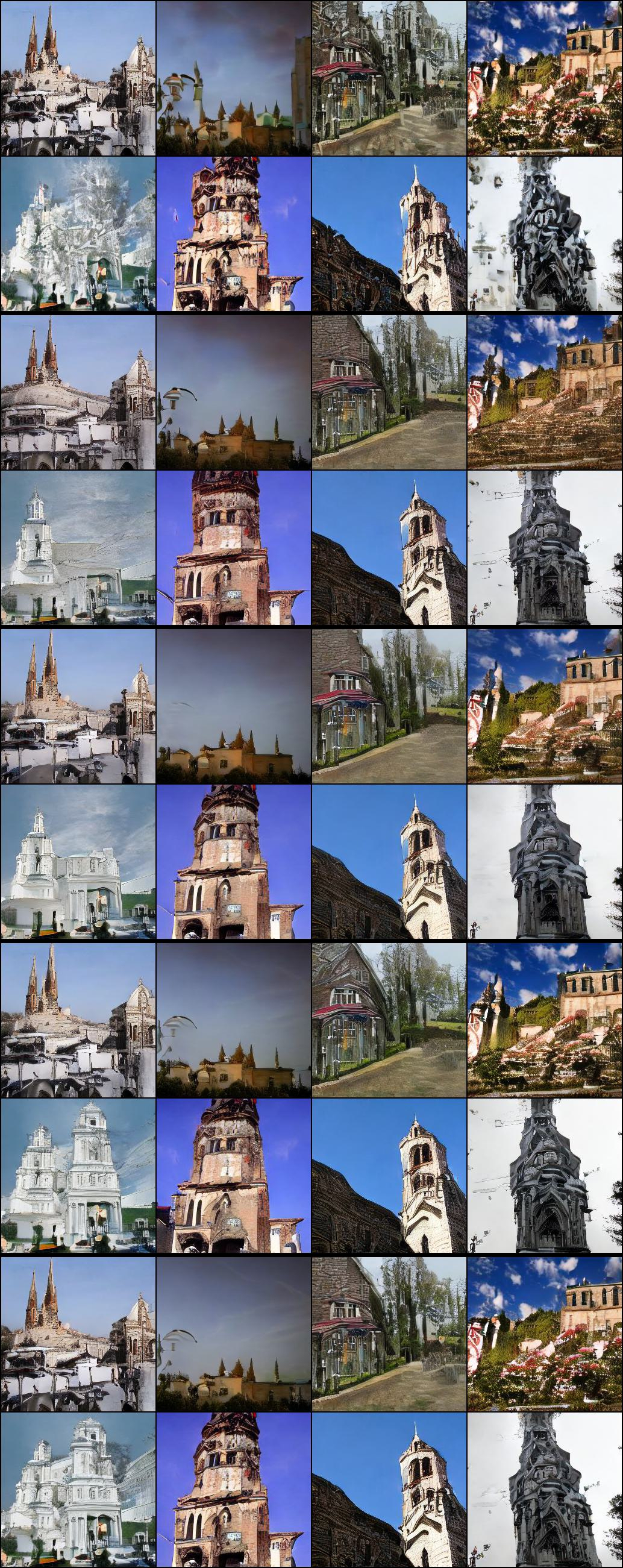}
    \caption{Quantization Learning Process of LSUN-Churches LDM-8}
    \label{fig:church-generation}
\end{wrapfigure}

EfficientDM used with 160 Learning Epochs to learn quantization mappings from FP to 4-bits weights

\vspace{5em}
\textbf{Epoch: 0}

\vspace{11.15em}
\textbf{Epoch: 40}

\vspace{11.18em}
\textbf{Epoch 80} 

\vspace{11.18em}
\textbf{Epoch: 120}

\vspace{11.18em}
\textbf{Epoch: 160}

\begin{figure}[t]
    \centering
    \vspace*{-1cm} 
    
    \textbf{  Ours (Mixed precision on PTQD) vs. PTQD Uniform} 
    
    \vspace{0.5cm} 
    
    \includegraphics[width=0.8\textwidth]{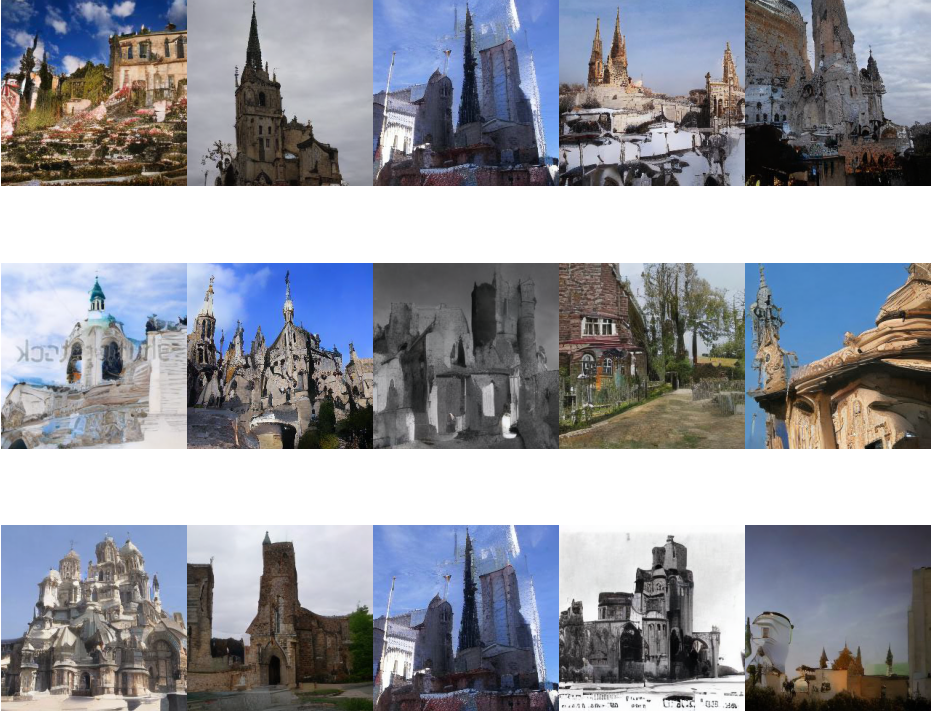}
    \caption{UNIFORM 4 bits -- 181 Mb, LSUN-Churches 256 × 256 LDM-8 (steps = 200, eta = 0.0)}
    \label{fig:churches_uniform_200}
    
    \vspace{0.8cm} 
    
    \includegraphics[width=0.8\textwidth]{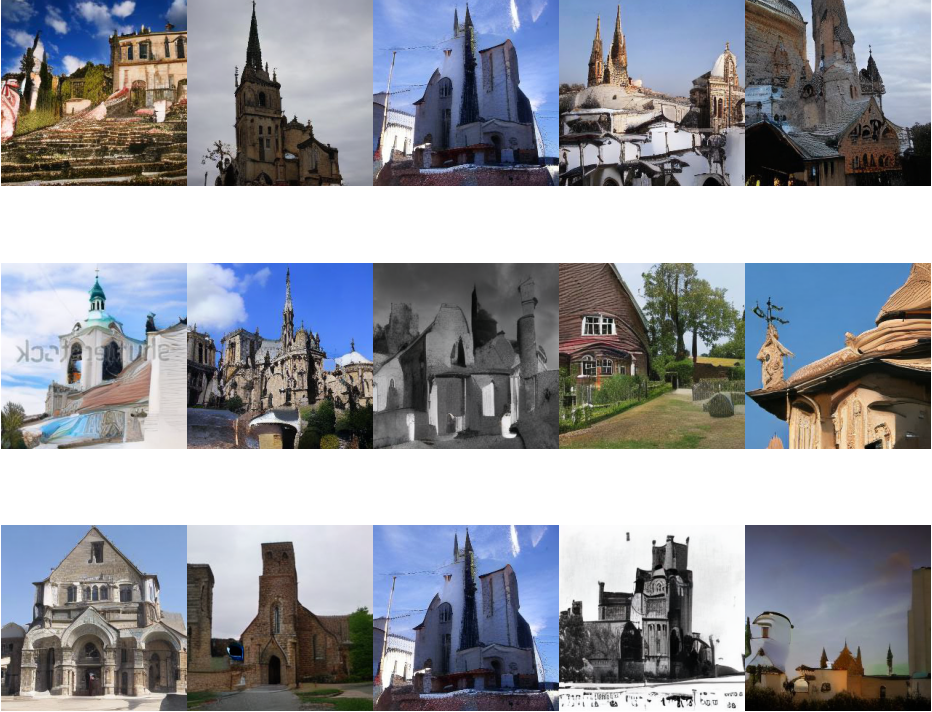}
    \caption{Mixed Precision -- 181 Mb, LSUN-Churches 256 × 256 LDM-8 (steps = 200, eta = 0.0)}
    \label{fig:churches_mixed_200}
\end{figure}

\begin{figure}[t]
    \centering
    \vspace*{-1cm} 
    
    \textbf{  Ours (Mixed precision on PTQD) vs. PTQD Uniform} 
    
    \vspace{0.5cm} 
    
    \includegraphics[width=0.8\textwidth]{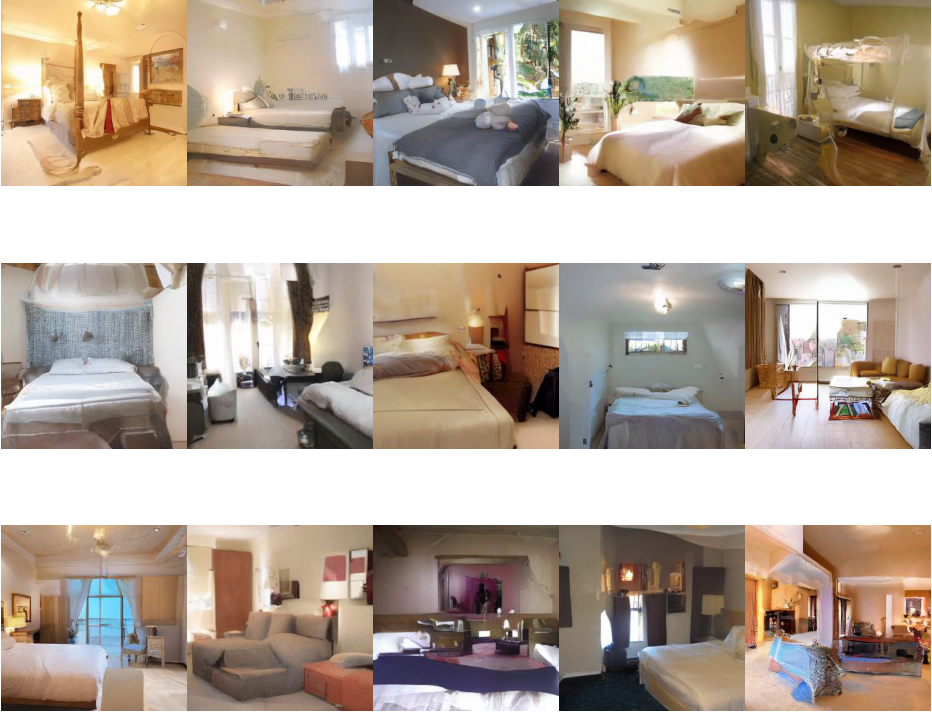}
    \caption{UNIFORM 4 bits -- 194 Mb, LSUN-Bedrooms LDM-4 (steps = 200, eta = 1.0)}
    \label{fig:bedrooms_uniform_200}
    
    \vspace{0.8cm} 
    
    \includegraphics[width=0.8\textwidth]{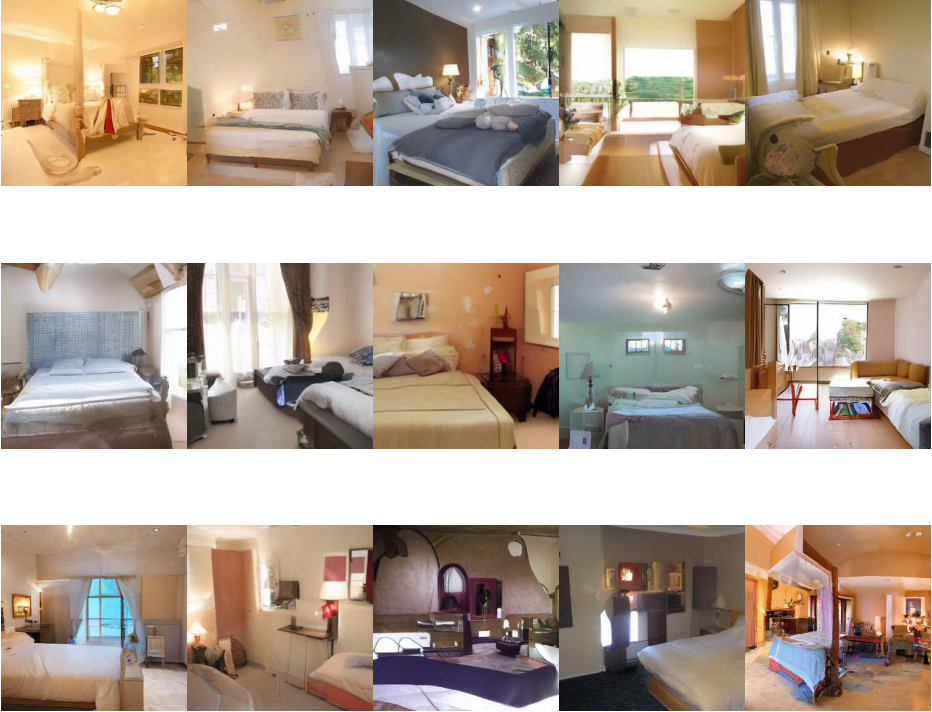}
    \caption{Mixed Precision -- 194 Mb, LSUN-Bedrooms LDM-4 (steps = 200, eta = 1.0)}
    \label{fig:bedrooms_mixed_200}
\end{figure}

\begin{figure}[t]
    \centering
    \vspace*{-1cm} 
    
    \textbf{  Ours (Mixed precision on PTQD) vs. PTQD Uniform} 
    
    \vspace{0.5cm} 
    
    \includegraphics[width=0.8\textwidth]{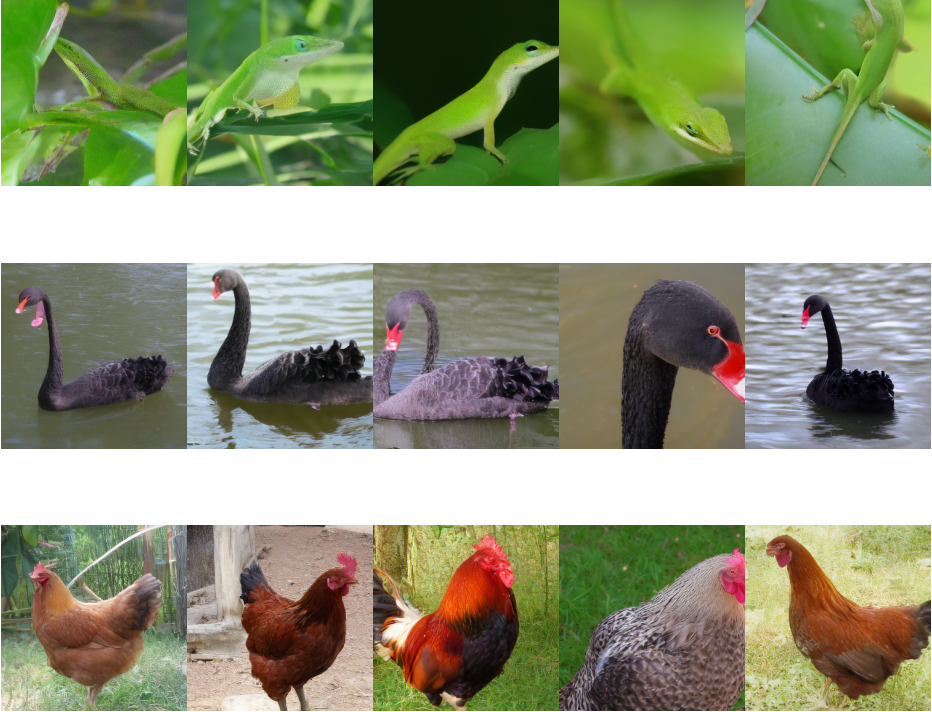}
    \caption{UNIFORM 4 bits -- 250 Mb, ImageNet LDM-4 (steps = 20, eta = 1.0, scale = 3.0)}
    \label{fig:imagenet_uniform}
    
    \vspace{0.8cm} 
    
    \includegraphics[width=0.8\textwidth]{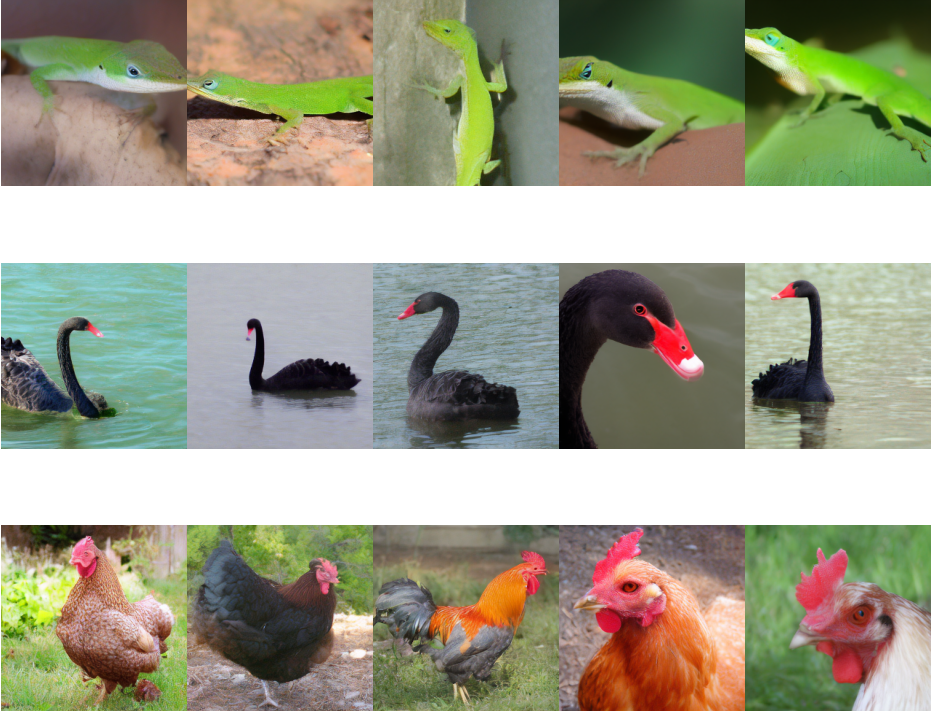}
    \caption{Mixed Precision -- 250 Mb, ImageNet LDM-4 (steps = 20, eta = 1.0, scale = 3.0)}
    \label{fig:imagenet_mixed}
\end{figure}

\begin{figure}[t]
    \centering
    \vspace*{-1cm} 
    
    \textbf{  Ours (Mixed precision on EfficientDM) vs. EfficientDM Uniform} 
    
    \vspace{0.5cm} 
    
    \includegraphics[width=0.8\textwidth]{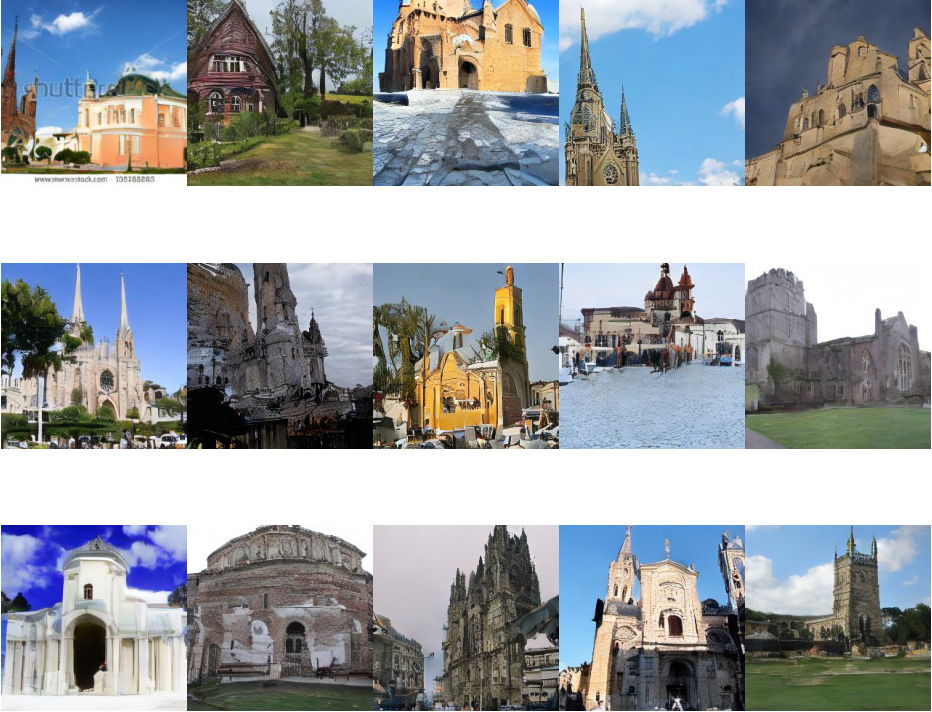}
    \caption{UNIFORM 4 bits -- 181 Mb, LSUN-Churches 256 × 256 LDM-8 (steps = 100, eta = 0.0)}
    \label{fig:churches_uniform}
    
    \vspace{0.8cm} 
    
    \includegraphics[width=0.8\textwidth]{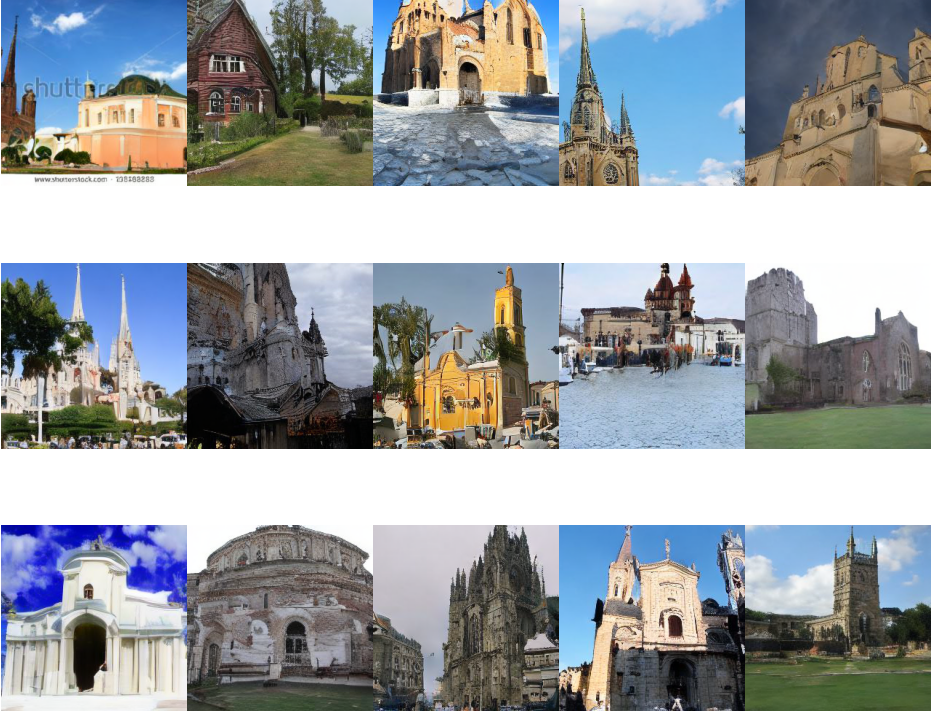}
    \caption{Mixed Precision -- 181 Mb, LSUN-Churches 256 × 256 LDM-8 (steps = 100, eta = 0.0)}
    \label{fig:churches_mixed}
\end{figure}

\begin{figure}[t]
    \centering
    \vspace*{-1cm} 
    
    \textbf{ Ours (Mixed precision on EfficientDM) vs. EfficientDM Uniform} 
    
    \vspace{0.5cm} 
    
    \includegraphics[width=0.8\textwidth]{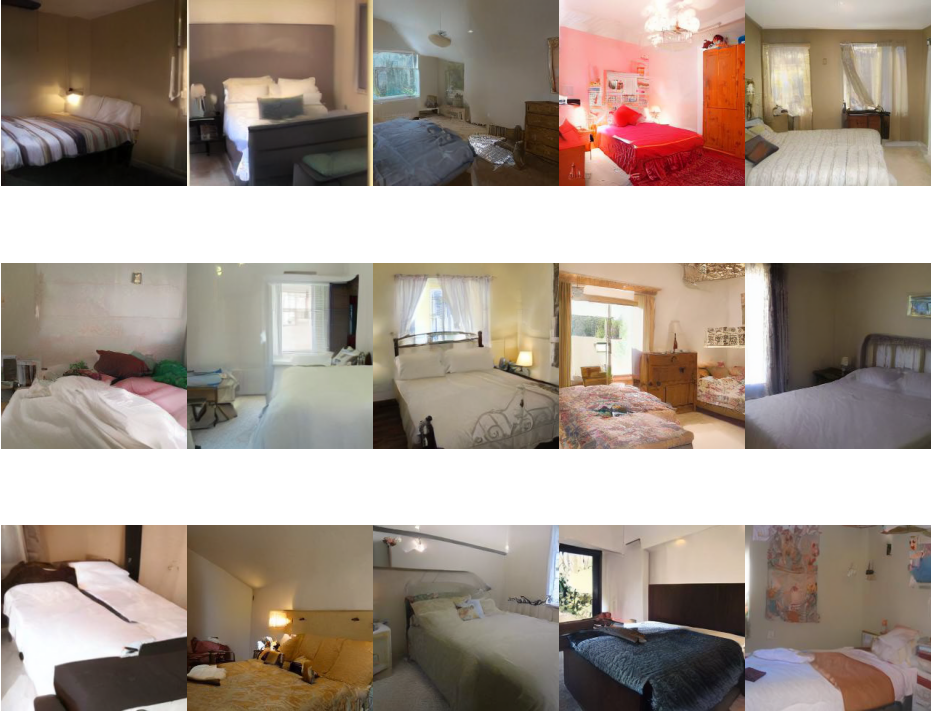}
    \caption{UNIFORM 4 bits -- 194 Mb, LSUN-Bedrooms LDM-4 (steps = 100, eta = 1.0)}
    \label{fig:beds_uniform}
    
    \vspace{0.8cm} 
    
    \includegraphics[width=0.8\textwidth]{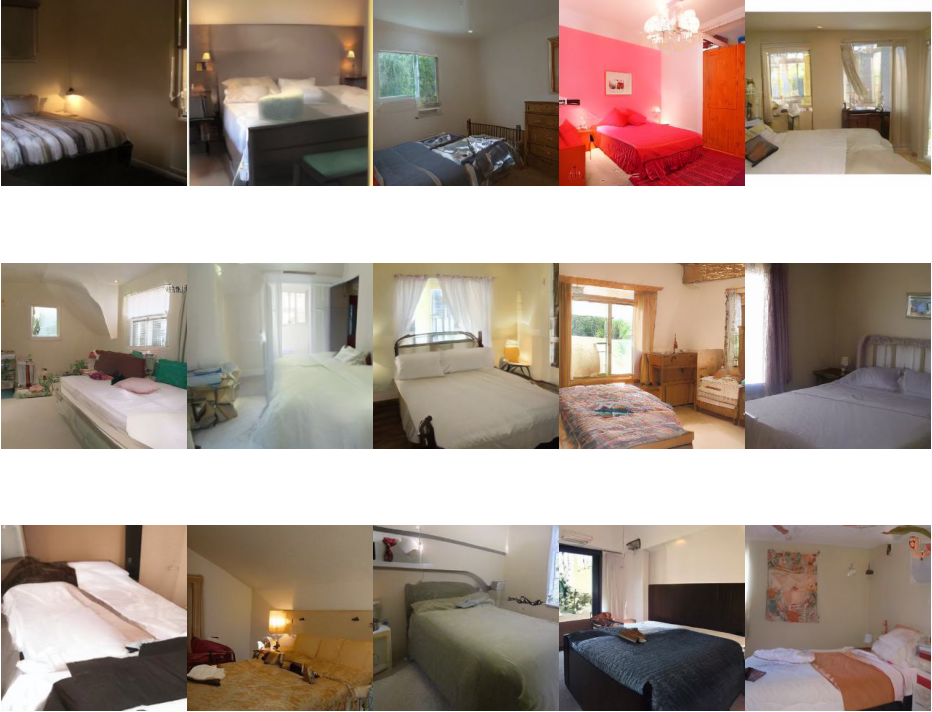}
    \caption{Mixed Precision -- 194 Mb, LSUN-Bedrooms LDM-4 (steps = 100, eta = 1.0)}
    \label{fig:beds_mixed}
\end{figure}

\begin{figure}[t]
    \centering
    \vspace*{-1cm} 
    
    \textbf{Ours (Mixed precision on EfficientDM) vs. EfficientDM Uniform} 
    
    \vspace{0.5cm} 
    
    \includegraphics[width=0.8\textwidth]{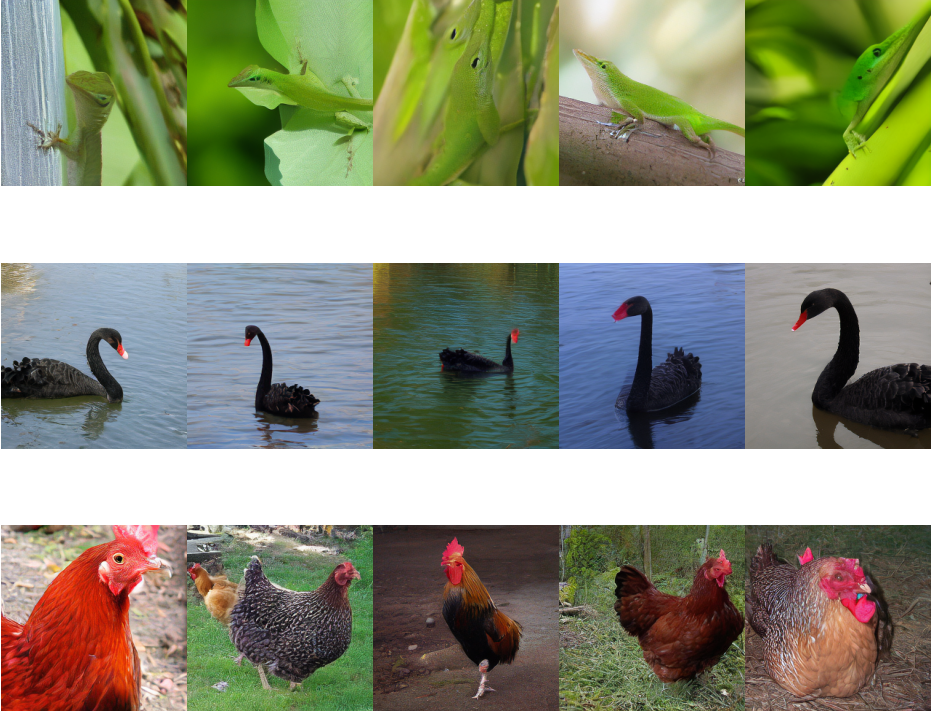}
    \caption{UNIFORM 4 bits -- 250 Mb, ImageNet LDM-4 (steps = 20, eta = 1.0, scale = 3.0)}
    \label{fig:image1}
    
    \vspace{0.8cm} 
    
    \includegraphics[width=0.8\textwidth]{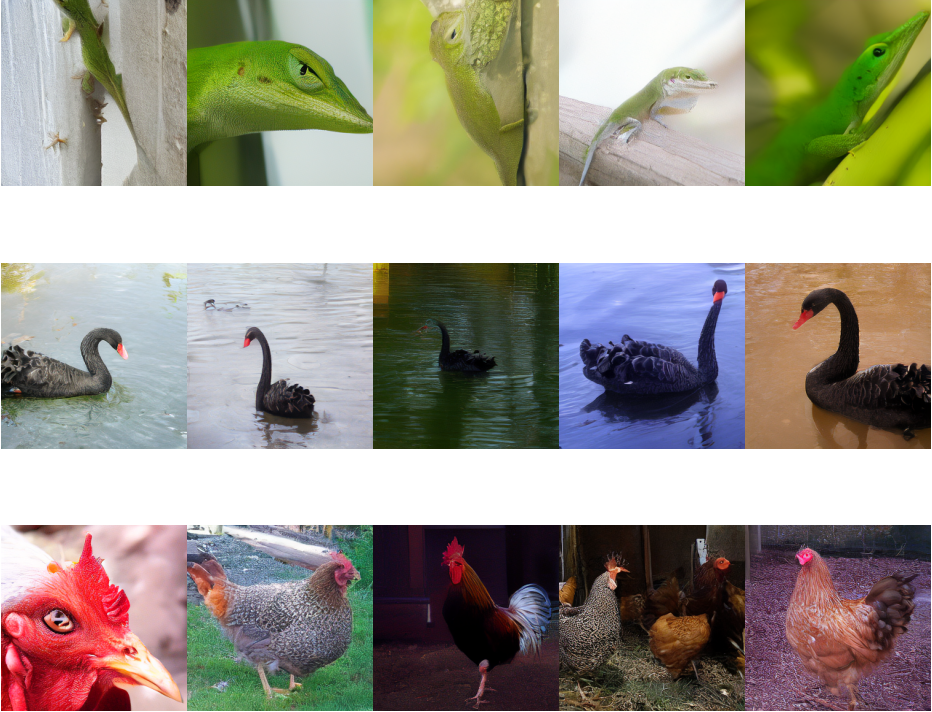}
    \caption{Mixed Precision -- 250 Mb, ImageNet LDM-4 (steps = 20, eta = 1.0, scale = 3.0)}
    \label{fig:image2}
\end{figure}

\begin{figure}
    \centering
    \includegraphics[height=0.95\textheight]{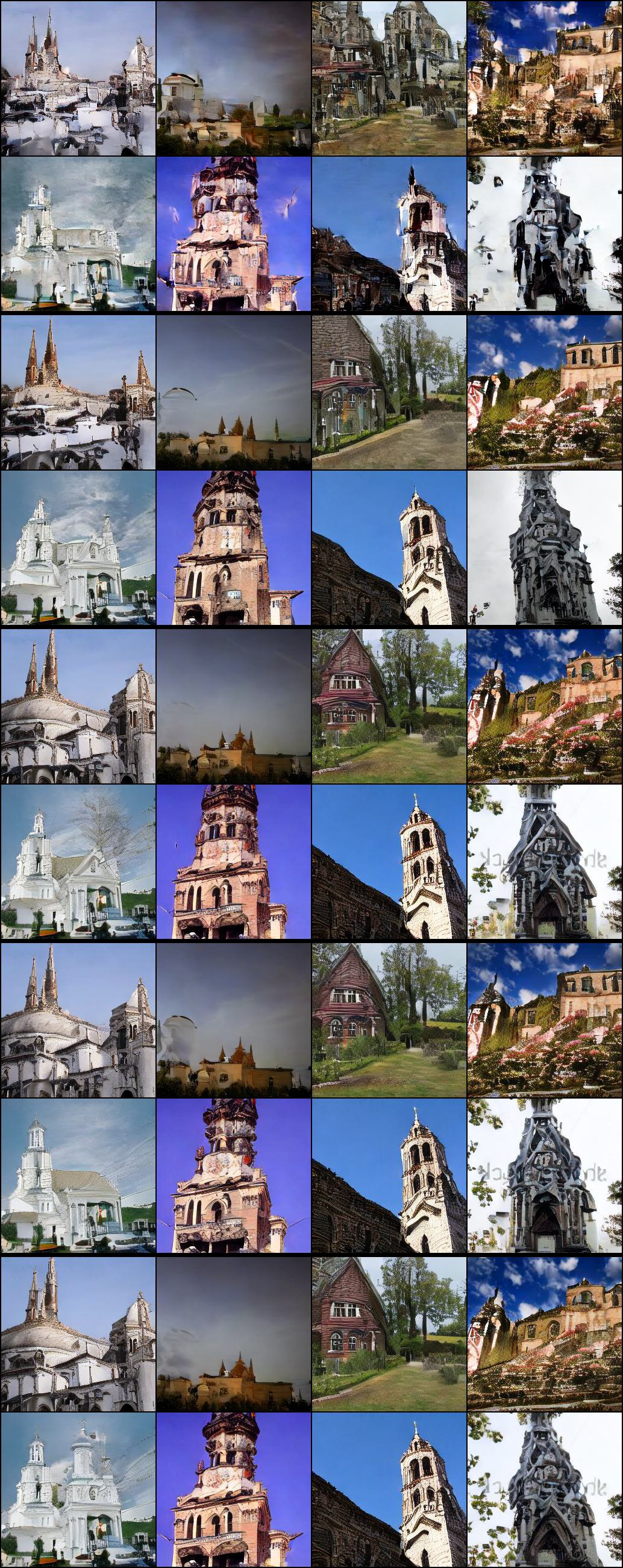}
    \caption{Comparison of images using differing sized models as in Figure 7, LSUN-Churches 256 × 256 LDM-8 (steps = 100, eta = 0.0) }
    \label{}
\end{figure}

\begin{figure}
    \centering
    \includegraphics[width=0.8\textwidth]{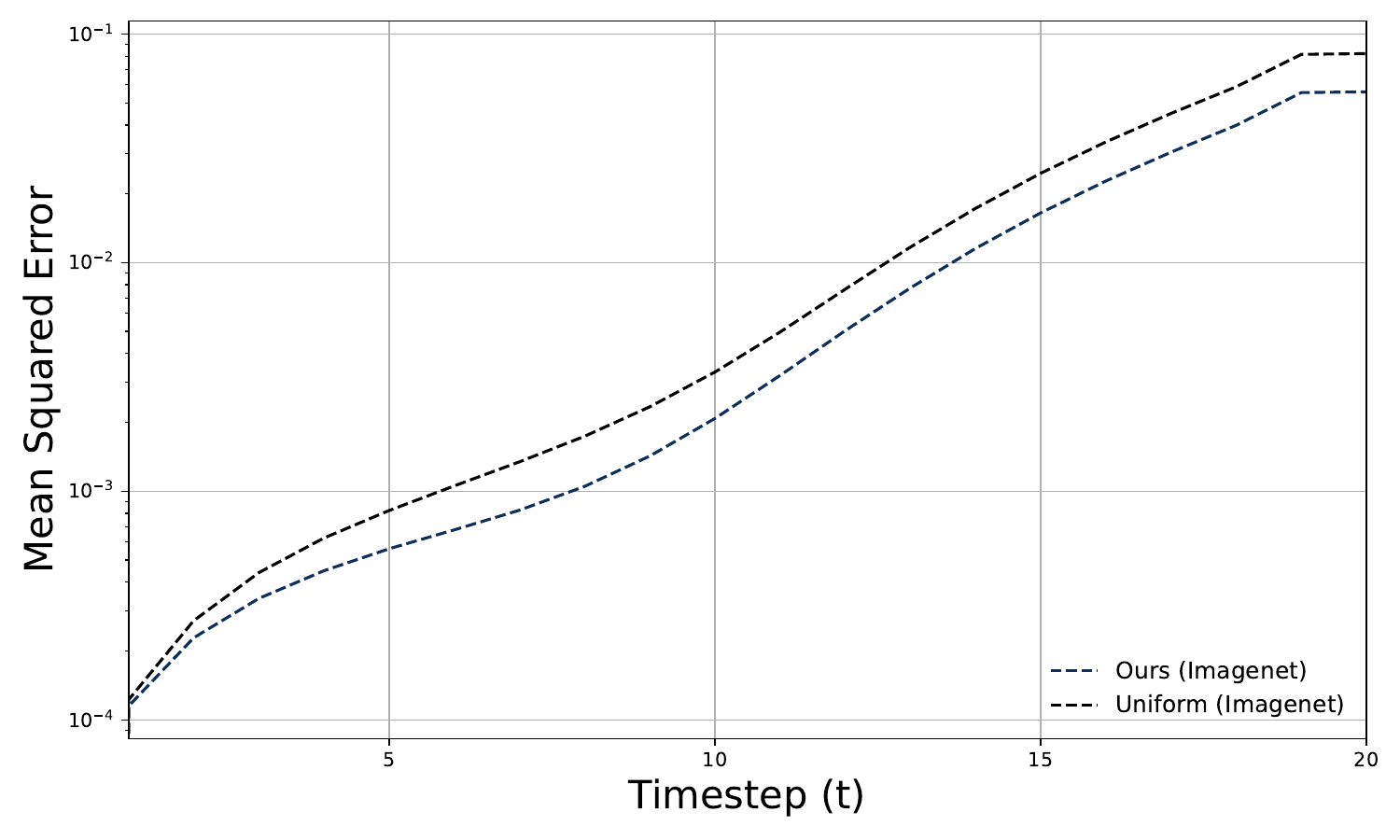}
    \caption{Mean square error between quantized and full precision image generation}
\end{figure}